\documentclass{article}
\usepackage{ijcai25}

\usepackage{times}
\usepackage{soul}
\usepackage{url}
\usepackage[hidelinks]{hyperref}
\usepackage[utf8]{inputenc}
\usepackage[small]{caption}
\usepackage{graphicx}
\usepackage{amsmath}
\usepackage{amsthm}
\usepackage{booktabs}
\usepackage{algorithm}
\usepackage{algorithmic}
\usepackage[switch]{lineno}

\usepackage{amssymb}

\title{GCNT: Graph-Based Transformer Policies\\for Morphology-Agnostic Reinforcement Learning}

\author{
Yingbo Luo$^1$\and
Meibao Yao$^1$\thanks{Corresponding Author}\And
Xueming Xiao$^2$\\
\affiliations
$^1$Jilin University\\
$^2$Changchun University of Science and Technology \\
\emails
luoyb23@mails.jlu.edu.cn,
meibaoyao@jlu.edu.cn,
alexcapshow@gmail.com
}

\begin{document}

\maketitle

\begin{abstract}

Training a universal controller for robots with different morphologies is a promising research trend, since it can significantly enhance the robustness and resilience of the robotic system. However, diverse morphologies can yield different dimensions of state space and action space, making it difficult to comply with traditional policy networks. Existing methods address this issue by modularizing the robot configuration, while do not adequately extract and utilize the overall morphological information, which has been proven crucial for training a universal controller. To this end, we propose GCNT, a morphology-agnostic policy network based on improved Graph Convolutional Network (GCN) and Transformer. It exploits the fact that GCN and Transformer can handle arbitrary number of modules to achieve compatibility with diverse morphologies. Our key insight is that the GCN is able to efficiently extract morphology information of robots, while Transformer ensures that it is fully utilized by allowing each node of the robot to communicate this information directly. Experimental results show that our method can generate resilient locomotion behaviors for robots with different configurations, including zero-shot generalization to robot morphologies not seen during training. In particular, GCNT achieved the best performance on 8 tasks in the 2 standard benchmarks.

\end{abstract}

\section{Introduction}

\begin{figure}[!htbp]
\centering
\includegraphics[width=1\columnwidth]{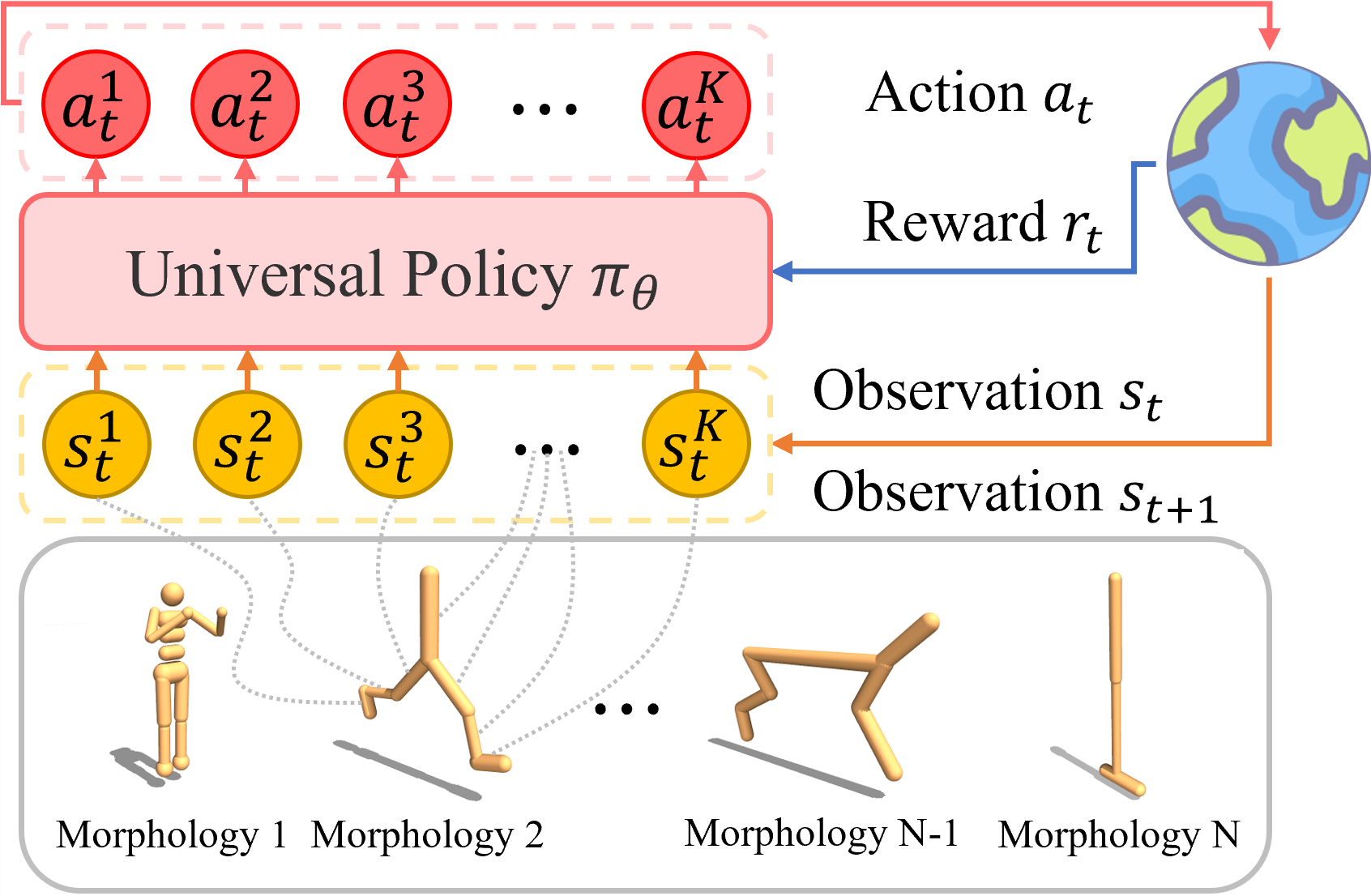}
\caption{Learning universal controllers with various morphologies.}
\label{fig1}
\end{figure}

In recent years, Deep Reinforcement Learning (DRL) has been widely used for continuous control tasks of mobile robots and has achieved satisfactory performance \cite{Mnih2015,Lillicrap2015,Hasselt2016,Fujimoto2018,Ye2020,xie2022mapless,sun2023co}. However, in traditional reinforcement learning, for each robot configuration, a specific policy network for control needs to be trained from scratch, that is, one network corresponds to one configuration. Since training each policy network requires substantial time and resources to interact with the environment, the overall cost is unacceptable when dealing with multiple robots. To address this issue, some scholars brought up the concept of the universal control network that allows all robot configurations to share a single policy network \cite{wang2018,Pathak2019,Huang2020,kurin2021,hong2022,chen2023}, that is, one policy network corresponds to multiple configurations. Such reinforcement learning models are called morphology-agnostic RL, inhomogeneous multi-task RL, or modular RL. Obviously, compared with traditional reinforcement learning, morphology-agnostic reinforcement learning has broader applications: First, it can significantly enhance the robustness and resilience of robots \cite{Zhang2017}. For example, a field exploration robot with a broken leg can still operate normally through the original policy network. Second, for unstructured terrain in the wild, we can focus on optimizing the configurations of robots that are best adapted to the current environment, without having to train a separate motion controller for each morphology \cite{wang2019}. Third, as in the fields of Natural Language Processing (NLP) \cite{Devlin2019,Brown2020} and Computer Vision (CV) \cite{Dosovitskiy2021,He2022,Han2022,tian20243d}, it can serve as a pre-trained prior for fine-tuning new morphologies \cite{gupta2022}. That is, for brand new robot configurations, we can use the parameters of the universal policy network as initialization and achieve satisfactory  control performance through a small number of iterative training. This saves a lot of training time and resources. Last, by providing a unified representation for robots of different morphologies, it makes training large models in the field of robotic control possible \cite{furuta2023}.

Since robots of different configurations have different numbers of joints and sensors, they have different state spaces and action spaces \cite{zhao2024dynainsremover,zhao2024m2cs}. However, the traditional MLP-based reinforcement learning model requires the input and output dimensions of the policy network be fixed, so they cannot control variably-configured robots. A natural idea is that any robot can be represented as a combination of multiple modules, and we can assume that the dimensions of the state and action space of each module are the same. If our policy network operates on a single module rather than the entire robot ( that is, the policy network receives local observations from individual modules as inputs and produces local actions for them as outputs), and the local actions cooperate to form the overall actions of the robot, then the problem that a single network cannot adapt to different robot configurations is solved.

However, there is a core problem that needs to be solved: how to better coordinate the discretized limbs of robots, so that the local actions of each module can be combined into the overall actions of the robot to accomplish various tasks. In other words, the key to solving the problem is how to determine the influence of the states of other modules on the current module. Previous studies generally opts for message passing mechanisms or Transformer architectures to address this issue. However, this approach is not ideal for the following two reasons. First, similar to the hidden state in Recurrent Neural Network (RNN) \cite{Hochreiter1997,Guo2020}, message passing mechanisms face the problem of limited capacity. After messages pass through multiple nodes, the original information contained in them is greatly diminished. This results in ineffective communication between nodes that are far apart (e.g., the hand and foot nodes of a humanoid robot). Second, the purely Transformer-based architecture neglects the robot's morphological information, which has been proven to be crucial for training a universal controller. Therefore, we propose GCNT, a morphology-agnostic robot control network architecture based on Graph Convolutional Network (GCN) \cite{Thomas2017} and Transformer. Our key insight is that GCN is able to efficiently extract morphology information of robots, while Transformer ensures that it is fully utilized by allowing each node of the robot to communicate this information directly. GCNT converts the robot's morphology into a corresponding graph, uses the local observations of individual modules as input, and generates local actions for each module as output, thereby achieving universal control for robots of different morphologies. In summary, our main contributions are as follows:
\begin{itemize}
\item We propose GCNT, a novel network architecture for morphology-agnostic reinforcement learning. It decomposes the problem into two independent processes: morphological information extraction and node communication, providing a new insight into robot control.

\item To fully extract the morphological information of the robot, we designed complementary GCN module and Weisfeiler-Lehman module in GCNT to overcome the indexing inconsistency of traversal-based methods. Besides, we improve the structure of GCN by adding additional linear layers and residual connections to avoid gradient vanishing caused by network depth increasing.

\item We conducted comparative experiments with 7 baselines in 2 standard benchmarks. Our method achieves leading performance in all 8 tasks and zero-shot generalization.
\end{itemize}

\section{Problem Formulation}

To enable the policy network to handle different state space and action space dimensions of various robots, we modularize the robots. In this manner we treat any morphology of a robot as a combination of multiple modules, where each module has identical state space and action space dimensions. The policy network takes the local state of each module as input and generates local actions for each module as output. Specifically, consider an agent with $K$ limbs, where we regard each limb and its corresponding joint (actuator) as a module. At each discrete time step $t$, the agent's state $s_t$ is composed of the local states $s^i_t$, that is, $s_t=\{s^1_t,s^2_t,\ldots,s^K_t\}$\, and the action output by the policy network is $a_t=\{a^1_t,a^2_t,\ldots,a^K_t\}$, where $a^i_t$ is the torque value for the corresponding actuator of the limb $k$. Note that for $\forall i,j\in\{1,2,\ldots,K\}$, we ensure $dim(s^i_t)=dim(s^j_t)$ and $dim(a^i_t)=dim(a^j_t)$.

To represent the relationship among modules, we use an undirected graph $\mathcal{G}:=\langle\mathcal{V}, \mathcal{E}\rangle$ to describe the robot's morphology. Each node $v_i\in\mathcal{V}$ for $i\in\{1,\ldots,K\}$ represents a limb of the agent, and each undirected edge $(v_i,v_j)\in\mathcal{E} $ indicates that limb $v_i$ and limb $v_j$ are directly connected through one or more joints. In practical applications, $\mathcal{E}$ can be represented by an adjacency matrix $A\in\{0,1\}^{K\times K}$, where $A^{i,j}=1$ if $v_i$ and $v_j$ are connected, otherwise $A^{i,j}=0$.

In this work, we assume there are $N$ agents with different morphologies in the configuration space. As shown in Figure 1, a universal policy network $\pi_\theta$ is trained to control them for locomotion, where $\theta$ represents the learnable parameters of the policy. We adopt a joint policy optimization approach to achieve morphology-agnostic reinforcement learning, which follows Markov Decision Process (MDP). For each discrete time step $t$, the policy network $\pi_\theta$ generates appropriate composite action $\{a^k_t\}^{K_n}_{k=1}$ based on the current state $\{s^k_t\}^{K_n}_{k=1}$ of agent $n$. After the agent $n$ executes the actions, the environment returns the next state $\{s^k_{t+1}\}^{K_n}_{k=1}$ and the corresponding reward $r^n_t$. Since the aim of our policy network is to perform well for all morphologies of robots, we need to consider the reward returns of all agents. Thus, our objective function is:
\begin{equation}
F(\theta)=\mathbb{E}_{\pi_\theta}\sum^{N}_{n=1}\sum^{\infty}_{t=0}\left[\gamma^tr^n_t\left(\{s^k_t\}^{K_n}_{k=1},\{a^k_t\}^{K_n}_{k=1}\right)\right] \label{eq1}
\end{equation}
where $a^k_t=\pi_\theta(s^k_t)$ and $\gamma$ is the discount factor. We aim to find the optimal solution $\theta^*$ that maximizes the expected cumulative return on discounts for all agents.
\begin{equation}
\theta^*=\arg\max_\theta F(\theta) \label{eq2}
\end{equation}

\begin{figure*}[!htbp]
\centering
\includegraphics[width=1\textwidth]{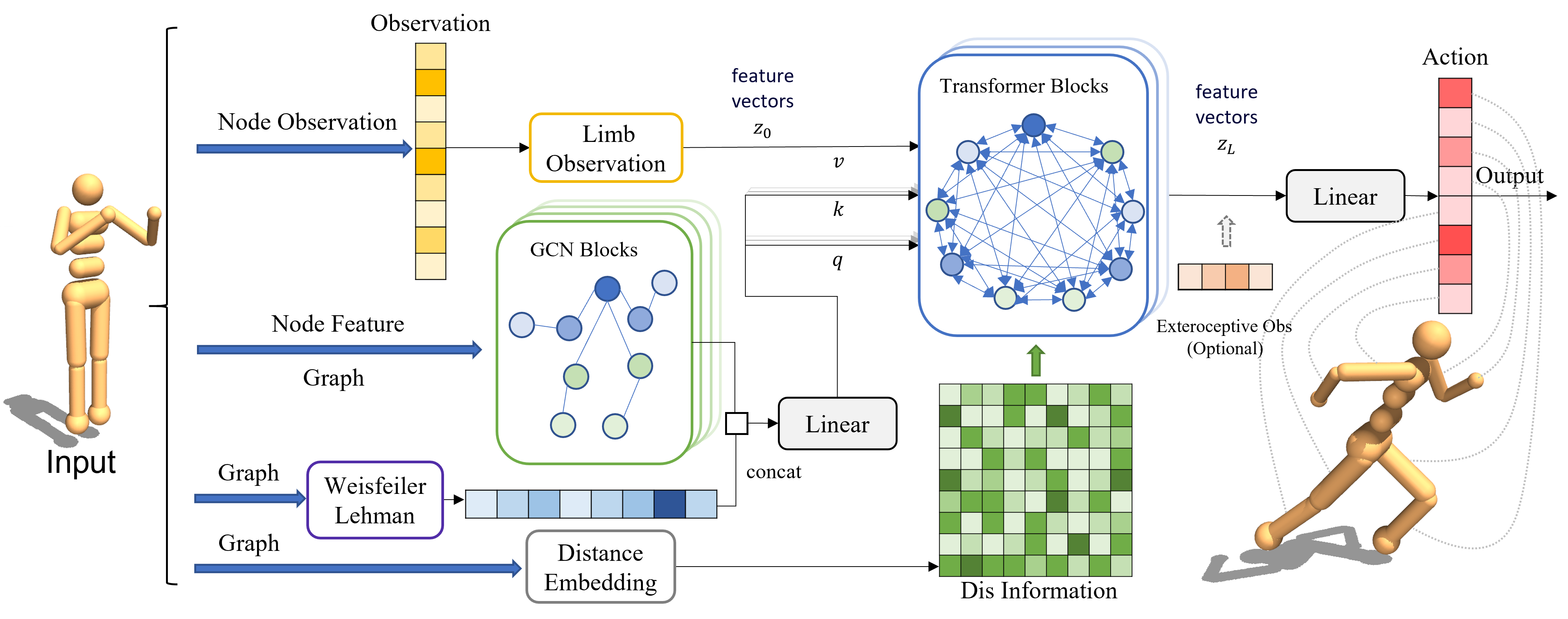} 
\caption{Overview of GCNT's network architecture, mainly composed of five modules: the limb observation module processes state information, the GCN and Weisfeiler-Lehman modules extract morphological information, the learnable distance embedding module considers the impact of physical distances among limbs, and the Transformer module is used for communication between limbs.}
\label{fig2}
\end{figure*}

\section{Methodology}

In this work, we present GCNT, a novel network architecture based on GCN and Transformer, for morphology-agnostic reinforcement learning in continuous control tasks. Our motivation is: GCN is able to aggregate information from neighboring nodes to the current node via graph convolution, which makes it naturally suitable for structural information extraction. Moreover, the Transformer allows each node to communicate directly and assess the importance of information through attention weights, thus preventing information loss during transmission. Therefore, we use these characteristics to solve the problem that discrete robot limbs cannot effectively cooperate with each other. More details of GCNT are shown below.

\subsection{Overview of GCNT}

The network architecture of GCNT is shown in Figure \ref{fig2}. For robots of any morphology, the input to GCNT can be divided into four parts, represented by the four blue arrows in Figure \ref{fig2}. They go through the limb observation module, the GCN module, the Weisfeiler-Lehman module, and the learnable distance embedding module, respectively. The limb observation module includes the current state information of each limb, which is the main basis for generating actions. The GCN module is proposed to extract the robot's local morphological information. The Weisfeiler-Lehman module is a complement to the GCN module and is used to extract the overall morphological information. The learnable distance embedding module determines the influence of limbs at different distances on the current limb. Afterwards, all information is aggregated in the Transformer module, and communication among limbs is completed through its attention mechanism. Finally, the results of communication are passed through a linear layer to obtain the actions that each limb should execute.

\subsection{Limb Observation Module}

This module is mainly used to process limb observations. As mentioned, the dimension of each limb's local observation are same. GCNT takes the local observation of each limb as input and generates local actions for each joint. In this work, the observation space of each limb in the two benchmarks remain the same as in previous work \cite{Huang2020,kurin2021,hong2022,gupta2022,Xiong2023}, which includes the motion state information and hardware attribute information of limbs and joints. The detailed information can be found in the Appendix D. All observations are uniformly passed through an MLP and mapped to a higher dimension for subsequent processing.

\begin{figure*}[!htbp]
\centering
\includegraphics[width=1\textwidth]{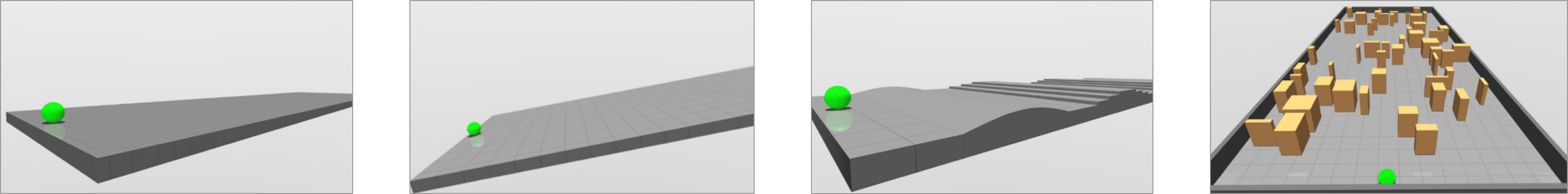}     
\caption{The four experiment environments in UNIMAL:  Flat Terrain (FT), Incline, Mixed Terrain (MT), and Obstacles. Images credit to Gupta et al. [2021; 2022]}
\label{fig3}
\end{figure*}

\subsection{GCN Module}

\begin{figure}[!htbp]
\centering
\includegraphics[width=0.8\columnwidth]{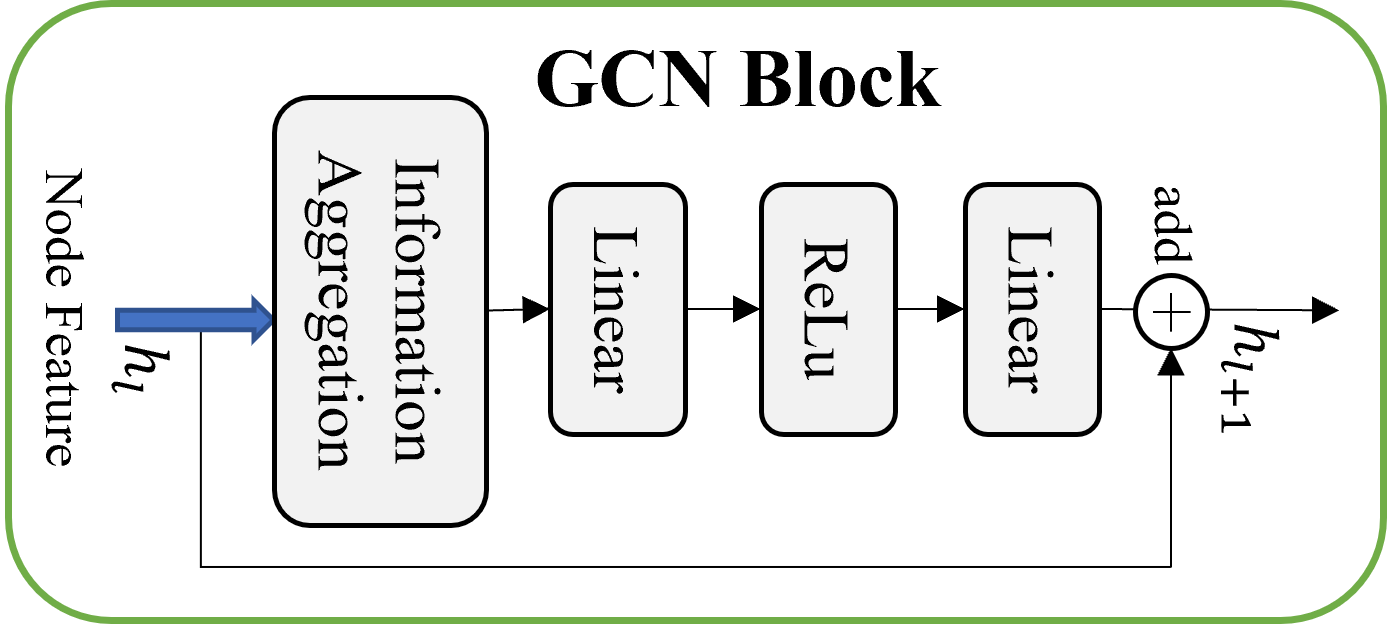}
\caption{Internal structure of the GCN block.}
\label{fig4}
\end{figure}
In this paper, we propose the improved GCN module to extract the local morphology information of the robot. The mathematical principle behind GCN is based on spectral graph theory. It uses the normalized graph Laplacian for feature aggregation and approximates the convolution operation on the graph as a polynomial filter. This updates each node's features as a weighted combination of its own features and those of its neighbors, so that the structure information in the graph can be captured effectively. The process of extracting information by GCN can be expressed as the following equation:
\begin{equation}
H^{(l+1)}=\sigma\left(\widetilde{D}^{-\frac{1}{2}} \widetilde{A} \widetilde{D}^{-\frac{1}{2}} H^{(l)} W^{(l)}\right) \label{eq3}
\end{equation}
where $\widetilde{A}=A+I_K$, $A$ is the adjacency matrix, and $I_K$ is the identity matrix of size $K$, $\widetilde D_{ii}= {\textstyle \sum_{j}}\widetilde A_{ij} $, $W^{(l)}$ is a layer-specific trainable weight matrix, $\sigma (\cdot)$ is an activation function, $H^{(l)}\in \mathbb{R}^{K\times D}$ is the feature matrix in the $l^{th}$ layer, and $H^{(0)}=X$. $X$ represents the original attributes of each node. In this work, we use one-hot encoding to represent different types of limbs and include them as the original attributes $X$ of each node in the GCN operations. Notably, we made changes to the structure of GCN, as shown in Figure \ref{fig4}. To enhance the extraction performance, we added additional linear layers to the network structure and we incorporated residual connections similar to those in ResNet \cite{He2016} to prevent vanishing gradient caused by the excessive depth of the GCNT network \cite{ying2021}. The computation process of the improved GCN is as follows:
\begin{equation}
H^{(l+1)}=\sigma\left(\widetilde{D}^{-\frac{1}{2}} \widetilde{A} \widetilde{D}^{-\frac{1}{2}} H^{(l)} W^{(l)}_1\right)W^{(l)}_2+H^{(l)} \label{eq4}
\end{equation}

\subsection{Weisfeiler-Lehman Module}

The Weisfeiler-Lehman (WL) algorithm is widely used for graph isomorphism testing and graph classification tasks \cite{weisfeiler1968}. In this work, we use the WL algorithm to capture the overall morphological information of the robots. We believe that the color string obtained through the WL algorithm can reflect the global structural information of the graph to a certain extent. The more similar the morphology of two robots is, the more similar the results obtained by the WL algorithm will be, and their control policy should also be more similar. Therefore, we use the WL module as a complement to the GCN module to capture the overall morphological information of the robot. In practical applications, we use the type of each limb as the initial color of each node and adopt a fixed number of iterations for all robot morphology. After the iteration ends, we use a vector to represent the occurrence times of each color during the iteration and use this vector as the overall morphological information, concatenating it with the local morphological information generated by the GCN module for each node.

\subsection{Learnable Distance Embedding Module \& Transformer Module}

\begin{figure}[!htbp]
\centering
\includegraphics[width=0.95\columnwidth]{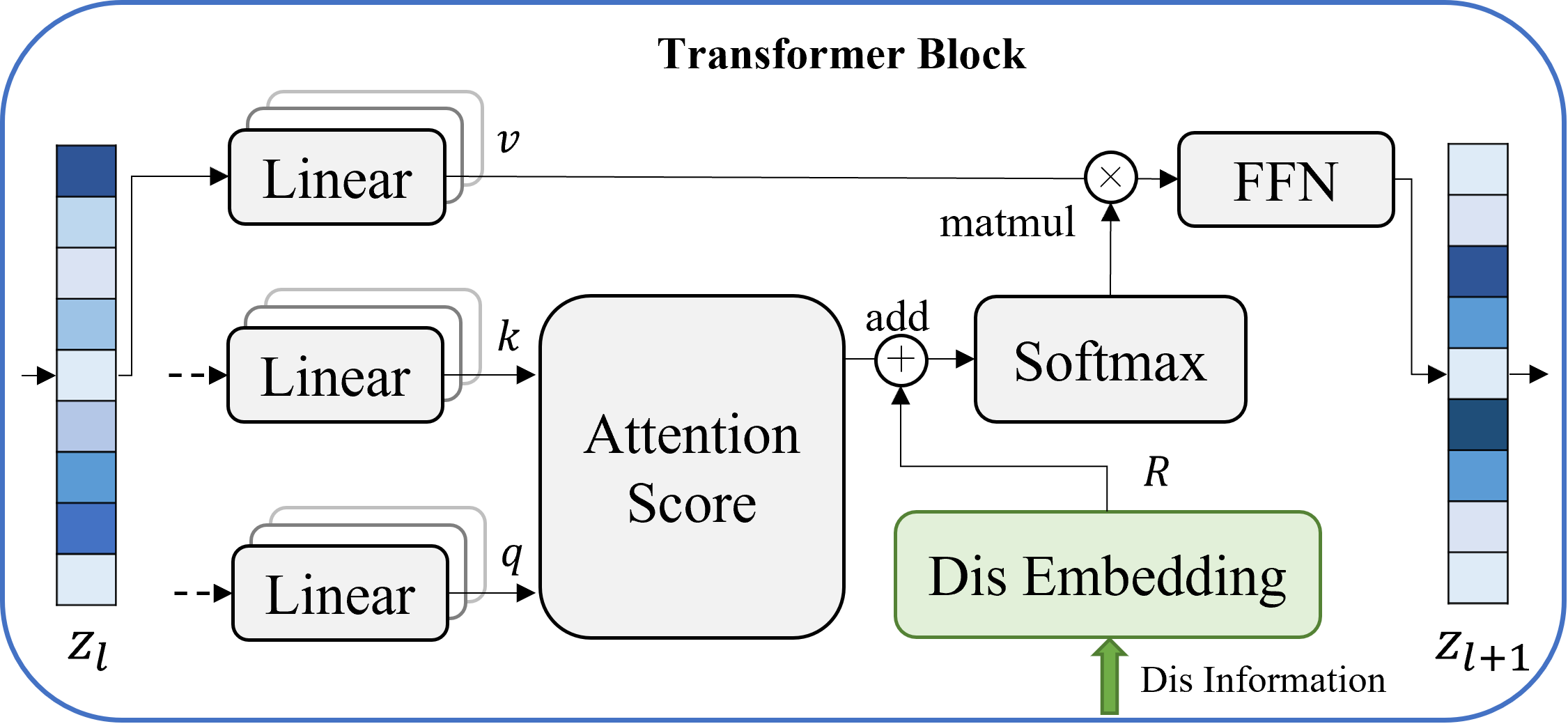}
\caption{Internal structure of the Transformer block.}
\label{fig5}
\end{figure}

The learnable distance embedding module is proposed to work with the Transformer module, considering the influence of distance in the communication between different limbs. For a given robot configuration represented by an undirected graph $\mathcal{G}=\langle\mathcal{V}, \mathcal{E}\rangle$, we denote the distance value of two directly connected nodes as 1 and use the Floyd algorithm to calculate the shortest distance $D$ among all nodes in the graph, as shown by the green matrix in Figure \ref{fig2}. We input this into the Transformer module, as shown in Figure \ref{fig5}. We propose a learnable distance embedding layer to map the limb distance information to each head of the multi-head attention layer. Specifically, $R^{i,j}=g_\phi(D^{i,j})$, where $g_\phi(\cdot)$ is a mapping function with parameters $\phi$, and $R^{i,j}$ is a vector with each element corresponding to a head in the multi-head attention. When calculating the attention weights, $R^{i,j}$ is added to the attention score and then passed through the Softmax function to reflect different impact of limbs at different distances on the current limb. This process can be expressed by Equation \ref{eq5}, where the subscript$(h)$ denotes the $h$-th head and $R^{i,j}_{(h)}$ is a $h$-th entry of $R^{i,j}$.
\begin{equation}
\alpha^{i,j}_{(h)}=\frac{q^i_{(h)}k^{jT}_{(h)}}{\sqrt{d_{(h)}}}+R^{i,j}_{(h)} \label{eq5}
\end{equation}

\subsection{Optimization Method}

In order to optimize the parameter $\theta$ in GCNT, we use an actor-critic architecture based on the deterministic policy gradient algorithm to find the optimal value of the objective function in Equation \ref{eq1}, which is a standard practice for continuous control tasks \cite{Lillicrap2015}. The actor and critic have the same network architecture, which remains the same as Figure \ref{fig2}. Specifically, we use the TD3 algorithm \cite{Fujimoto2018} in benchmark 1 and the PPO algorithm \cite{John2017} in benchmark 2 to ensure a fair comparison with previous works.

\begin{figure*}[!htbp]
\centering
\includegraphics[width=1\textwidth]{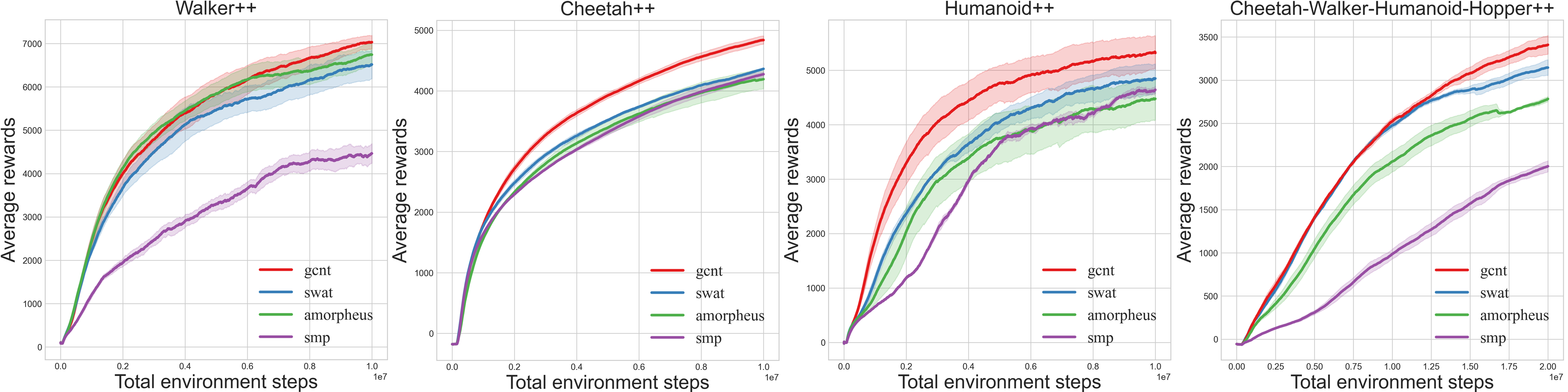}     
\caption{Average rewards across all agents in environments of Walker++, Cheetah++, Humanoid++, CWHH++ in SMPENV.}
\label{fig6}
\end{figure*}

\begin{figure*}[!htbp]
\centering
\includegraphics[width=1\textwidth]{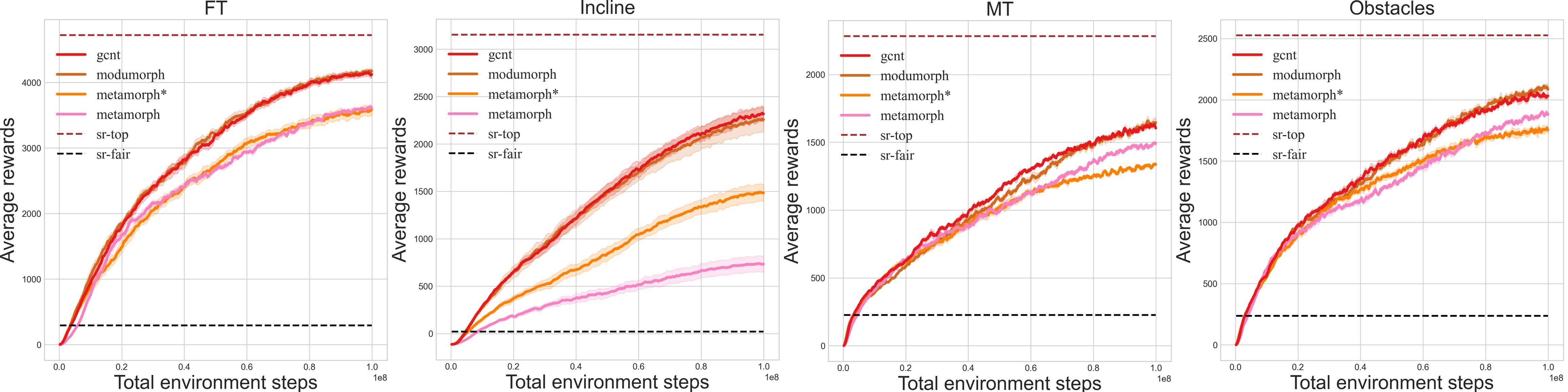}     
\caption{Average rewards across all agents in environments of FT, Incline, MT and Obstacles in UNIMAL.}
\label{fig7}
\end{figure*}

\section{Experimental Results}

We conducted experiments across 8 different scenarios on two standard benchmarks and compared with 7 baselines, ensuring thorough testing. The benchmarks are SMPENV \cite{Huang2020} and UNIMAL \cite{gupta2021}. All tests are based on the MuJoCo \cite{Todorov2012}.

In SMPENV, we set up four test environments: (1) Walker++, (2) Cheetah++, (3) Humanoid++, (4) Cheetah-Walker-Humanoid-Hopper++ (CWHH++). These four environments can be divided into two categories, in-domain and cross-domain. For the first three test environments, robots of different morphologies in the environment have some similarities. Each test environment contains complete morphologies and variants with missing limbs of the same category of robots. For example, Humanoid++ includes both the complete humanoid\_2d\_9\_full and the humanoid\_2d\_7\_left\_leg, which is missing the left leg. While the last CWHH++ environment is a comprehensive test environment. The morphologies and the movement manner of the robots in the environment vary significantly. For example, the walker needs to move by alternately coordinating its left and right legs, while the hopper can only move by jumping with a single leg.

In UNIMAL, we conducted experiments in four environments. As shown in Figure \ref{fig3}, they are: (1) Flat Terrain (FT), (2) Incline, (3) Mixed Terrain (MT), and (4) Obstacles. In all environments, the goal of the robot is to pursue the farthest possible travel distance. 
Additionally, the terrain in MT and the pillars in Obstacles are randomized in each round. So the last two environments provide a height map centered around the robot as an extra external observation input, allowing the agent to perceive and respond to different terrains or obstacles. Finally, it is worth noting that in UNIMAL we need to control 100 robots with different morphologies. So UNIMAL has greater complexity than SMPENV.

\subsection{Training Results in SMPENV}

To demonstrate the effectiveness of our method, GCNT, we compare it with three baselines: the message passing method SMP \cite{Huang2020}, the Transformer-based method AMORPHEUS \cite{kurin2021}, and the Structure-aware Transformer method SWAT \cite{hong2022}. We run all experiments with 3 random seeds to report the mean and the standard error. For fairness, we use the TD3 algorithm to train the policy network for all methods. The results are shown in Figure \ref{fig6}, where the solid line represents the mean and the shaded area represents the standard error.

It can be observed that our method, GCNT, has the highest sample efficiency and achieves the best performance in all test environments. The experimental results validate our analysis. SMP, which uses the message-passing mechanism, suffers from severe information loss during multi-hop transmission, making it difficult for two distant limbs to coordinate. Thus it performed the worst in all environments. AMORPHEUS did not utilize morphological information. While it performed well for simpler robot configurations like Walker++ (each robot consisting of only a torso and two legs), it showed poor performance for more complex robot configurations. SWAT incorporated morphological information into the Transformer, but the information extracted based on traversal and graph methods does not always have a positive effect. For example, in the Walker++ environment, SWAT's performance is inferior to that of AMORPHEUS, which uses a pure Transformer architecture. This indicates that the morphological information extracted by SWAT actually hinders the training of the policy network. In contrast, the morphological information extracted by our GCNT is useful in all scenarios. Moreover, GCNT consistently outperforms SWAT, demonstrating that our method has a stronger ability to extract morphological information.

\subsection{Training Results in UNIMAL}

In UNIMAL, we compare the method with four baselines: the depth-first traversal-based method MetaMorph \cite{gupta2022}, the variant MetaMorph* \cite{Xiong2023}, the hypernetwork-based method ModuMorph \cite{Xiong2023}, and the methods SR-fair and SR-top, which train a separate policy network for each robot. Both SR-fair and SR-top use an MLP with 3 hidden layers and 256 hidden units in each layer. The difference is that SR-fair trains each robot with the same budget as used in multi-robot training (1M steps), while SR-top trains each robot until convergence (10M steps). SR-top can be viewed as an approximate upper bound. All methods use the PPO algorithm for training (using 4 seeds) and the experimental results are shown in Figure \ref{fig7}. As can be seen, all methods exhibit higher sample efficiency than SR-fair, and our method, GCNT, achieves leading performance across all scenarios. It is worth noting that although ModuMorph has similar effects to our method, the number of parameters of ModuMorph is 2-3 times that of our method because it uses a Hypernetwork to generate different encoders and decoders for each limb. More importantly, it has worse generalization, as detailed in the zero-shot experiments. Additionally, our method can also integrate Hypernetworks to generate different encoders and decoders for each body part to enhance performance. The corresponding experimental results can be found in the Appendix G. 

\begin{table}[!htbp]
\begin{center}
\begin{tabular}{lllll}
\hline
                         & \textbf{gcnt}              & \textbf{swat}      & \textbf{amorpheus}         & \textbf{smp}              \\ \hline
\textbf{w3}       & 171$\pm$ 4             & 417 $\pm$  34  & \textbf{651 $\pm$  36} & 142 $\pm$  6          \\
\textbf{w6}       & \textbf{2933$\pm$ 116} & 935 $\pm$  22  & 1463 $\pm$  110        & 1424 $\pm$  76        \\
\textbf{h7}     & \textbf{3509$\pm$ 9}   & 2464 $\pm$  60 & 883 $\pm$  53          & 1056 $\pm$  27        \\
\textbf{h8}     & \textbf{3332$\pm$ 27}  & 958 $\pm$  34  & 1598   $\pm$  88       & 304 $\pm$  4          \\
\textbf{c3}      & 96$\pm$ 8              & -157 $\pm$  2  & 87 $\pm$  6            & \textbf{211 $\pm$  8} \\
\textbf{c5}      & \textbf{2440$\pm$ 18}  & 986 $\pm$  12  & 2245 $\pm$  17         & 2149 $\pm$  7         \\
\textbf{c6}      & \textbf{4780$\pm$ 21}  & 3252 $\pm$  7  & 4322 $\pm$  9          & 4726 $\pm$  8         \\ \hline
\end{tabular}
\caption{Zero-shot generalization to unseen morphologies. The w, h, and c in the first column are abbreviations for walker, humanoid, and cheetah, respectively.}
\label{tab1}
\end{center}
\end{table}

\subsection{Zero-Shot Generalization to Unseen Morphologies}

The goal of learning a universal controller is not limited to training on specified morphologies but also includes zero-shot generalization to new morphologies without any further fine-tuning. Table \ref{tab1} shows the results of the zero-shot generalization to unseen morphologies in SMPENV. Each method was trained on the training sets of Walker++, Humanoid++, and Cheetah++ and evaluated on their corresponding test sets. We evaluated the average performance and the standard error over 3 random seeds, with each seed evaluated on 100 rollouts. It can be seen that GCNT achieved the highest average score in 5 out of the 7 tests. The cases where GCNT's generalization was not as good as other methods occurred in simpler configurations with fewer limbs (walker\_3\_main and cheetah\_3\_balanced both have only 3 limbs). However, as the number of limbs increased, GCNT’s performance improved. This further validates that GCNT can effectively extract and utilize morphological information of robots. To sum up, we believe that GCNT's performance in zero-shot generalization is superior to other methods.

\begin{figure}[!htbp]
\centering
\includegraphics[width=1\columnwidth]{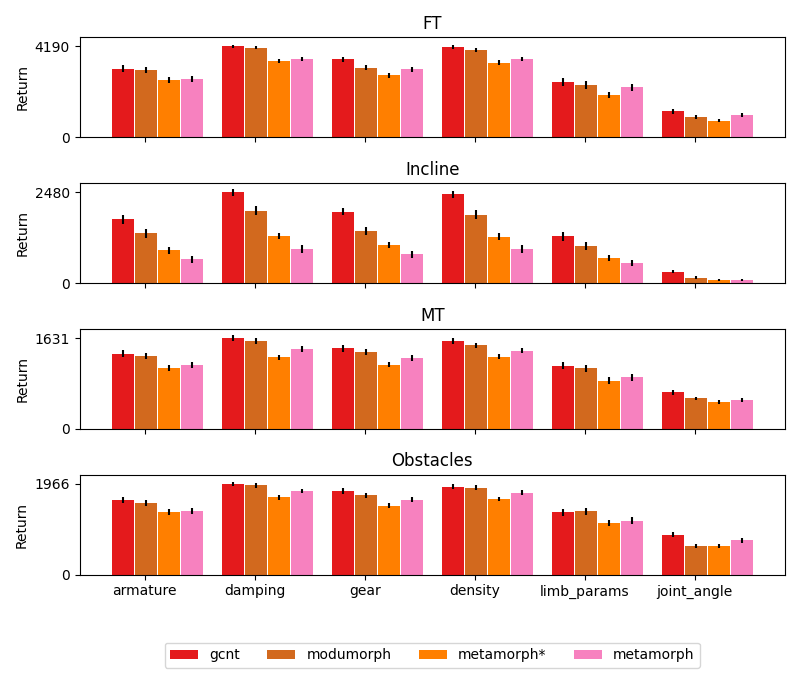}
\caption{Zero-shot generalization to kinematics and dynamics variations. Error bars denote 95\% bootstrapped confidence interval.}
\label{fig8}
\end{figure}

\subsection{Zero-Shot Generalization to Kinematics and Dynamics Variations}

We also conducted generalization experiments on dynamics and kinematics in UNIMAL. In this test set, the topological structure of each robot remains the same as in the training set, while the attributes of the limbs and joints are changed (including armature, density, damping, gear, module shape, and joint angle). We performed generalization experiments on 400 new robots, and the corresponding results are shown in Figure \ref{fig8}. As can be seen, our method, GCNT, exhibits the strongest generalization ability to both kinematics and dynamics.

\subsection{Ablation Study}

\begin{figure}[!htbp]
\centering
\includegraphics[width=0.7\columnwidth]{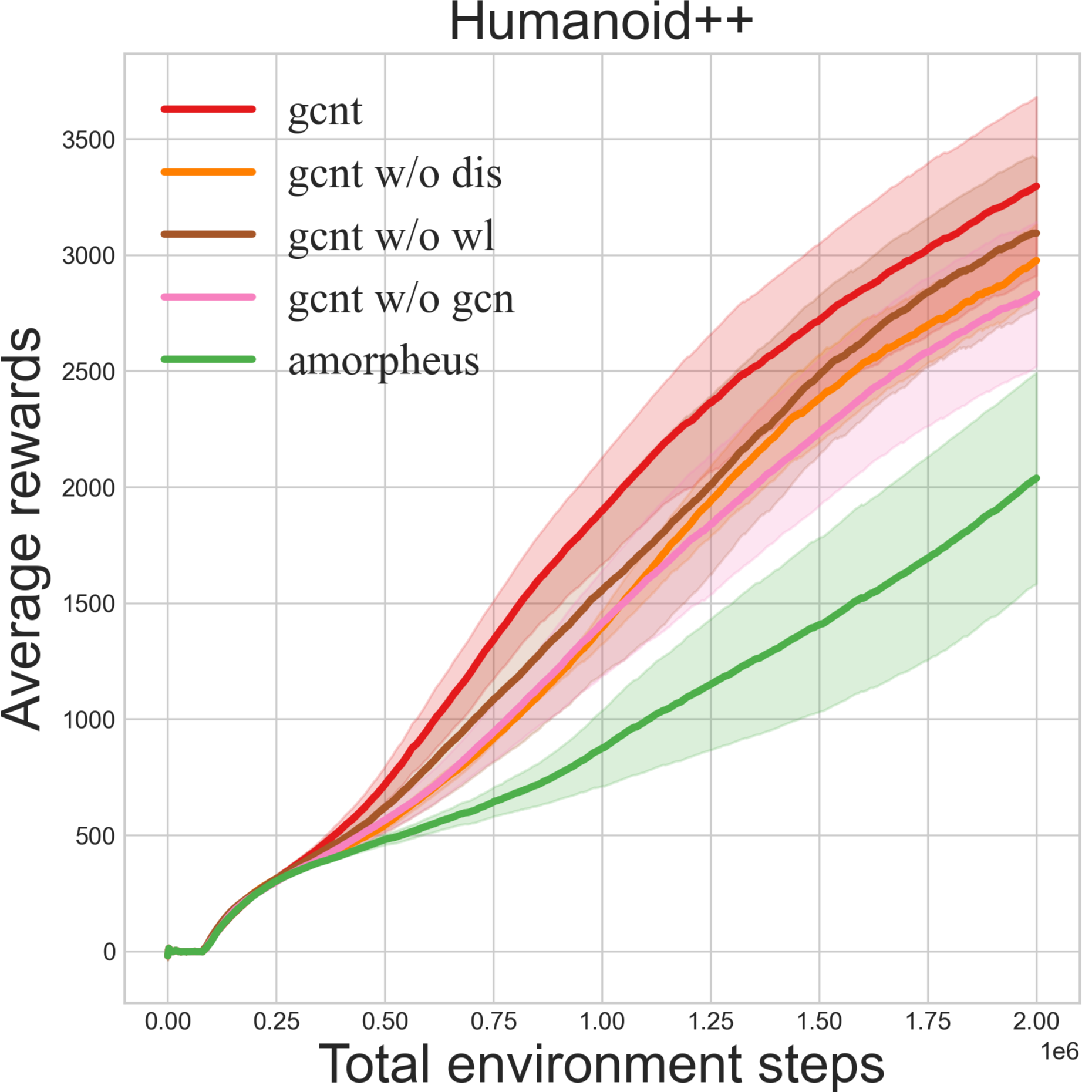}
\caption{The results of ablation study of GCNT.}
\label{fig9}
\end{figure}

To demonstrate the role of each module in GCNT, we conducted the ablation study. Specifically, we removed the GCN module, Weisfeiler-Lehman module, and learnable distance embedding module separately. All methods were evaluated on Humanoid++ because it has the most limbs and the most complex configuration, providing a complete test for validating the effectiveness of each module. The specific experimental results are shown in Figure \ref{fig9}. It can be seen that removing each module led to a performance drop, but the network still outperformed the pure Transformer AMORPHEUS.

\subsection{Morphological Information Embedding}

\begin{figure}[!htbp]
\centering
\includegraphics[width=1.0\columnwidth]{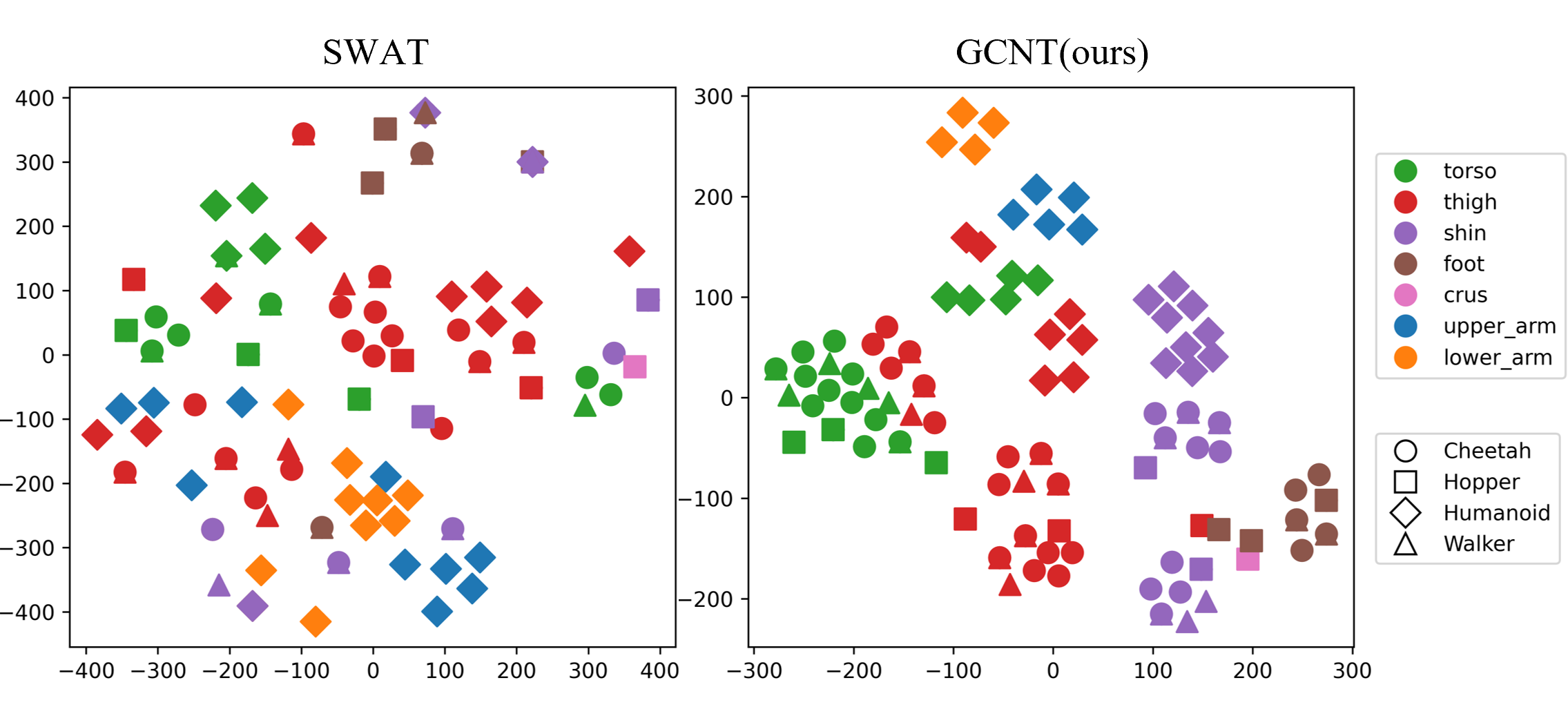}
\caption{Morphological information embedding learned by GCNT and SWAT from different robot configurations.}
\label{fig10}
\end{figure}

Previous work has shown that the morphological information of a robot is crucial for training a universal controller. However, previous methods use depth-first traversal for extraction \cite{hong2022,gupta2022}, which does not guarantee that limbs of the same type have the same index in different robots, as shown in the Appendix H. To observe whether GCNT extracts the information effectively, we visualize the structural information learned in the CWHH++ environment. Specifically, we use the t-SNE algorithm \cite{Maaten2008} to map the 128-dimensional structural vectors extracted by GCNT and SWAT into 2 dimensions. The corresponding results are shown in Figures \ref{fig10}, where the color of the points represents the type of limb, and the shape of the points represents the robot configuration to which it belongs. It can be seen that for GCNT, limbs with similar functions in different robot formation are mapped closely, while for SWAT, limbs with similar functions in different robot formation are mapped very scattered. This shows that SWAT's traversal-based method has different performance on different robot configurations and cannot effectively extract the robot's structural information. In contrast, the method based on GCN blocks and Weisfeiler-Lehman blocks in GCNT is effective for all robot configurations. This also explains why GCNT performs better in locomotion tasks and zero-shot generalization experiments.

\section{Related Work}

In general, research on morphology-agnostic reinforcement learning can be categorize as three stages. In the first stage, the core of related work is the use of message-passing mechanisms. \citeauthor{wang2018} [\citeyear{wang2018}] proposed NerveNet, which is based on Graph Neural Network (GNN) \cite{sanchez2021}. \citeauthor{Pathak2019} [\citeyear{Pathak2019}] and \citeauthor{Huang2020} [\citeyear{Huang2020}]  converted the robot's configurations into corresponding graphs, and proposed Dynamic Graph Network (DGN) and Shared Modularization Policy (SMP) as the policy networks shared among robot modules. In the second stage, the focus of research shifted to the utilization of Transformers. To address the issue of critical information loss during multi-hop communication in message-passing mechanisms. \citeauthor{kurin2021} [\citeyear{kurin2021}] propose a Transformer \cite{Vaswani2017} policy network AMORPHEUS. It utilizes a multi-head self-attention mechanism, allowing direct communication among robot modules without the need for multiple hops through different nodes. Notably, the work by \citeauthor{kurin2021} [\citeyear{kurin2021}] initiated a trend in research on Transformer-based policy networks, and most subsequent work has been based on this architecture. In the third stage, researchers began to integrate the robot's morphological information into the Transformer. \citeauthor{hong2022} [\citeyear{hong2022}] proposed SWAT, a Transformer-based policy network that incorporates traversal-based positional embeddings in node attributes and graph-based relational embeddings in the multi-head attention mechanism to encode this information.
\citeauthor{gupta2022} [\citeyear{gupta2022}] and \citeauthor{Xiong2023} [\citeyear{Xiong2023}] tried to use morphological information on larger robot benchmarks and proposed MetaMorph and ModuMorph respectively. \citeauthor{patel2024getzero} [\citeyear{patel2024getzero}] leveraged embodied graph connectivity for structural bias in attention mechanism.

Although the aforementioned research has made some progress in morphology-agnostic reinforcement learning tasks, there are still many shortcomings. For instance, it does not ensure efficient communication between nodes or adequately extract the morphological information of robots. These are areas worth further exploration in the future.

\section{Conclusion}

In this work, we propose GCNT, a modular network architecture based on GCN and Transformer, for morphology-agnostic reinforcement learning. Through the effective extraction of robot morphological information by the GCN module and the direct communication among different limbs by the Transformer module, GCNT can provide resilient locomotion control for robots of any morphology. Compared to previous work, our method achieves leading performance in all eight environments and zero-shot generalization experiments under Mujoco platform.

\section*{Acknowledgments}

This work was sponsored by the National Natural Science Foundation of China (NSFC) through grant No. 52472448, and Deep Earth Probe and Mineral Resources Exploration-National Science and Technology Major Project through grants No. 2024ZD1000800  and No. 2024ZD1000804. We thank the above mentioned funds for their financial support.

\bibliographystyle{named}
\bibliography{ijcai25}

\appendix
\onecolumn
\Large

\begin{center}
{\fontsize{18}{22}\selectfont\textbf{Appendix}}
\end{center}
\section{Implementation Details}

For the two standard benchmarks used for testing, we implemented GCNT on top of the SWAT and ModuMorph shared codebases, respectively. We modified their structure to make it compatible with our proposed module. Specifically, we concatenated the output of the Weisfeiler-Lehman module to each vector generated by the GCN module to obtain feature vectors containing morphological information. Then, these feature vectors are mapped to higher dimensions and add to the vectors generated by the limb observation module. Finally, the result of this addition is fed into the Transformer module, where feature fusion and processing are carried out under the impact of distance embedding information. It is worth noting that the residual and normalization layers in Transformer are not shown in the figure of the full paper due to space limitations. Additionally, we used the residual connections that span the entire network as reported in \cite{kurin2021}, connecting the residual states of each limb's observation to the output of the Transformer module to prevent nodes from forgetting their features after passing through each module.

For the SMPENV benchmark, we used the TD3 algorithm to learn the policy network. Initially, the robot randomly samples actions from a normal distribution $\mathcal{N}\left(\mathbf{a}_t, \sigma^2\right)$ to explore the environment, where $a_t$ is the mean action and the variance $\sigma^2$ is a hyperparameter used to control the randomness of the TD3 exploration policy. Then after the exploration phase, the robot uses the learned policy network to generate actions for each joint. During training, we set up a separate replay buffer for each agent morphology. All agents interact with the environment sequentially to collect samples. Once all agents have completed one episode, training is conducted for each agent in turn, and this process is repeated. For the Walker++, Cheetah++, and Humanoid++ test environments, we trained for 10 million environment steps. For the more complex CWHH++ environment, which has 32 different morphological agents, we trained for 20 million environment steps. Our model ran on a single NVIDIA A40 GPU, requiring seven days of training to reach 10 million environment steps.

For the UNIMAL benchmark, we use the PPO algorithm to learn the policy network. The PPO algorithm is more stable compared to the TD3 algorithm and is less sensitive to hyperparameter choices. During the training process, all morphological agents share the same buffer and use a dynamic replay buffer balancing scheme based on episode length to avoid 'rich gets richer' [Gupta et al., 2022]. For each test environment, we train with 100 million environment steps. It is worth noting that UNIMAL provides a more efficient algorithm implementation, and our model requires only 1.5 days of training on a single NVIDIA A40 GPU to reach 100 million environment steps.\\

\section{Hyperparameters}

In this section, we present all the hyperparameters used for GCNT. The hyperparameters vary in different benchmarks and these hyperparameters can be classified as related to the GCNT network architecture and optimization algorithms. The specific values for each hyperparameter are shown in Tables \ref{a.tab1}-\ref{a.tab5}.

\begin{table}[htbp!]
\Large
\centering
\begin{tabular}{lr}
\hline
\textbf{Hyperparameter Name}       & \textbf{Hyperparameter Value} \\ \hline
Normalization                      & LayerNorm      \\
GCN layers                         & 4              \\
GCN Activation                     & relu           \\
GCN output size                    & 16             \\
Transformer layers                 & 3              \\ 
Transformer heads                  & 2              \\ 
Transformer Feedforward dimension  & 256            \\
Transformer activation             & relu           \\  
Transformer input size             & 128            \\ \hline
\end{tabular}
\captionsetup{font=Large}
\caption{GCNT hyperparameter settings in SMPENV.}
\label{a.tab1}
\end{table}

\newpage
In SMPENV, it is worth noting that we kept the hyperparameters related to the TD3 algorithm the same across the four methods, SMP, AMORPHEUS, SWAT, and GCNT, to eliminate irrelevant factors and allow for better performance comparison. The detailed values are shown in the Table \ref{a.tab2}.

\begin{table}[htbp!]
\Large
\centering
\begin{tabular}{lr}
\hline
\textbf{Hyperparameter Name}      & \textbf{Hyperparameter Value} \\ \hline
Num of Random Seeds                  & 3                             \\ 
Mini-batch Size                   & 100                           \\ 
Replay Buffer Size                & 500000                        \\ 
Max Episode Length                & 1000                          \\ 
Max Replay Size Total             & 4000000                       \\ 
In-domain Environment Steps       & 10000000                      \\ 
Cross-domain Environment Steps    & 20000000                      \\ 
Policy Update Interval            & 2                             \\ 
Initial Exploration Steps         & 10000                         \\ 
Policy Noise                      & 0.2                           \\ 
Policy Noise Clip                 & 0.5                           \\ 
Target Network Update Rate $\tau$ & 0.046                         \\ 
Std of Gaussian Noise $\sigma$    & 0.126                         \\ 
Discount Factor                   & 0.99                          \\ 
Gradient Clipping                 & 0.1                           \\ 
Learning Rate                     & 0.0001                        \\ \hline
\end{tabular}
\captionsetup{font=Large}
\caption{Hyperparameter settings for TD3 in SMPENV.}
\label{a.tab2}
\end{table}

\begin{table}[htbp!]
\Large
\centering
\begin{tabular}{lr}
\hline
\textbf{Hyperparameter Name}       & \textbf{Hyperparameter Value} \\ \hline
Normalization                      & LayerNorm      \\
GCN layers                         & 4              \\
GCN activation                     & relu           \\
GCN output size                    & 23             \\
Transformer layers                 & 5              \\ 
Transformer heads                  & 2              \\ 
Transformer Feedforward dimension  & 1024           \\
Transformer activation             & relu           \\  
Transformer input size             & 128            \\ \hline
\end{tabular}
\captionsetup{font=Large}
\caption{GCNT hyperparameter settings in UNIMAL.}
\label{a.tab3}
\end{table}

\newpage

Similarly, in UNIMAL, we have kept the hyperparameters associated with the PPO algorithm the same in all methods. The detailed parameter values are shown below.

\begin{table}[htbp!]
\Large
\centering
\begin{tabular}{lr}
\hline
\textbf{Hyperparameter Name} & \textbf{Hyperparameter Value} \\ \hline
Num of Random Seeds & 4 \\ 
Discount $\gamma$ & 0.99 \\ 
GAE Parameter $\lambda$ & 0.95 \\ 
Policy Epochs & 8 \\ 
Batch Size & 5120 \\ 
Number of Parallel Environments & 32 \\ 
Total Timesteps & $1 \times 10^8$ \\ 
Optimizer & Adam \\ 
Initial Learning Rate & 0.0003 \\ 
Learning Rate Schedule & Linear warmup and cosine decay \\ 
Warmup Iterations & 5 \\ 
Gradient Clipping ($l_2$ norm) & 0.5 \\ 
Value Loss Coefficient & 0.2 \\ \hline
\end{tabular}
\captionsetup{font=Large}
\caption{Hyperparameter settings for PPO in UNIMAL.}
\label{a.tab4}
\end{table}

As mentioned in Gupta et al. [2022], the early stopping threshold $\delta$ is a key hyperparameter in PPO, and it has a significant impact on the final training results. Therefore, we performed a grid search over the candidate set \{0.03, 0.05\} and reported the optimal $\delta$ value for each method in each environment in Table 5.

\begin{table}[htbp!]
\Large
\centering
\begin{tabular}{ccccc}
\hline
\textbf{Environment} & \textbf{MetaMorph*} & \textbf{ModuMorph} & \textbf{GCNT} \\ \hline
FT        & 0.05 & 0.05 & 0.05  \\
INCLINE   & 0.03 & 0.05 & 0.05  \\
VT        & 0.03 & 0.03 & 0.03  \\
OBSTACLES & 0.03 & 0.03 & 0.03  \\ \hline
\end{tabular}
\captionsetup{font=Large}
\caption{Optimal value of the early stopping threshold for each method.}
\label{a.tab5}
\end{table}

\section{Details of the Environment}

Table \ref{a.tab6} shows the specific training and test sets for each environment in the zero-shot generalization experiment in SMPENV. For each environment, we first completed the training of the policy network on the training set, and then transferred the trained policy network to the test set for evaluation.

\begin{table}[!h]
\Large
\centering
\begin{tabular}{lll}
\hline
\textbf{Environment}          & \textbf{Train set}                                       & \textbf{Test set}                          \\ \hline
\textbf{Hopper++}             & hopper\_3                                                &                                            \\ 
                              & hopper\_4                                                &                                            \\ 
                              & hopper\_5                                                &                                            \\ \hline
\textbf{Walker++}             & walker\_2\_main                                          & walker\_3\_main                            \\ 
                              & walker\_4\_main                                          & walker\_6\_main                            \\ 
                              & walker\_5\_main                                          &                                            \\ 
                              & walker\_7\_main                                          &                                            \\ \hline
\textbf{Cheetah++}            & cheetah\_2\_back                                         & cheetah\_3\_balanced                       \\ 
                              & cheetah\_2\_front                                        & cheetah\_5\_back                           \\ 
                              & cheetah\_3\_back                                         & cheetah\_6\_front                          \\ 
                              & cheetah\_3\_front                                        &                                            \\ 
                              & cheetah\_4\_allback                                      &                                            \\ 
                              & cheetah\_4\_allfront                                     &                                            \\ 
                              & cheetah\_4\_back                                         &                                            \\ 
                              & cheetah\_4\_front                                        &                                            \\ 
                              & cheetah\_5\_balanced                                     &                                            \\ 
                              & cheetah\_5\_front                                        &                                            \\ 
                              & cheetah\_6\_back                                         &                                            \\ 
                              & cheetah\_7\_full                                         &                                            \\ \hline
\textbf{Humanoid++}           & humanoid\_2d\_7\_left\_arm                               & humanoid\_2d\_7\_left\_leg                 \\ 
                              & humanoid\_2d\_7\_lower\_arms                             & humanoid\_2d\_8\_right\_knee               \\ 
                              & humanoid\_2d\_7\_right\_arm                              &                                            \\ 
                              & humanoid\_2d\_7\_right\_leg                              &                                            \\ 
                              & humanoid\_2d\_8\_left\_knee                              &                                            \\ 
                              & humanoid\_2d\_9\_full                                    &                                            \\ \hline
\textbf{CWHH++}               & Union of                                                 & Union of                                   \\
                              & Hopper++,                                                & Walker++                                   \\
                              & Walker++,                                                & Cheetah++                                  \\
                              & Cheetah++,                                               & Humanoid++.                                \\
                              & Humanoid++.                                              &                                            \\ \hline

\end{tabular}
\captionsetup{font=Large}
\caption{Full list of environments in SMPENV.}
\label{a.tab6}
\end{table}

\newpage
\section{Details of the Observation}

In SMPENV, specific observations for each limb include limb type (e.g., hip or shoulder), position with a relative x-coordinate of the limb with respect to the torso, linear and angular velocities, joint angle, and possible range of the angle values normalized to [0, 1].

In UNIMAL, the observations of each module can be divided into four parts. (1) limb-related: relative x-coordinate position to the torso, linear and angular velocities, pose, inverse kinematics, inverse quaternion, attitude quaternion. (2) limb hardware-related: mass, shape. (3) joint-related: normalized angles, velocities, motion axes (4) joint hardware-related: range of motion, rotation axis information, actuator gear ratio.\\

\section{Details of the Joint Optimization}

As mentioned, in order to be consistent with previous work, we used different optimization methods in different benchmarks. The details of the optimization using the TD3 and PPO algorithms are shown in Algorithm \ref{a.alg1} and Algorithm \ref{a.alg2}, respectively.

\begin{algorithm}
    \Large
    \captionsetup{font=Large}
    \caption{Joint Training of All Agents by TD3}
    \label{a.alg1}
    \textbf{Symbols and Notations}:\\
    $s^{(n)}_t$: all limbs' states of agent $n$ at time $t$\\
    $a^{(n)}_t$: all limbs' actions of agent $n$ at time $t$\\
    $r^{(n)}_t$: rewards of agent $n$ at time $t$\\
    $done^{n}$: whether agent $n$ is done\\
    $rb^{n}$: replay buffer for agent $n$\\
    $T_{episode}$: maximum time step of an episode
    \begin{algorithmic}[1]
    \STATE initialize \textit{GCNT} with random parameter $\theta$
    \STATE empty replay buffer $rb^{n}$ for each agent $n$
    \WHILE{not reach the maximum time step}
        \FORALL{agent $n$}
            \FOR {t = 0, \dots, $T_{episode}$} 
				\STATE $a^{(n)}_t$ $\gets$ \textit{GCNT}($s^{(n)}_t$)
				\STATE $s^{(n)}_{t+1},r^{(n)}_t,done^{n}$ $\gets$ simulate($e,a^{(n)}_t$)
				\STATE add \{$s^{(n)}_t,s^{(n)}_{t+1},a^{(n)}_t,r^{(n)}_t,done^{n}$\} to $rb^{n}$
                \IF {$done^n$}
                    \STATE break
                \ENDIF
			\ENDFOR
		\ENDFOR
		\FORALL{agent $n$}
			\STATE \textit{GCNT} $\gets$ train with TD3(\textit{GCNT}, $rb^{n}$)
		\ENDFOR
    \ENDWHILE
    \end{algorithmic}
\end{algorithm}

\begin{algorithm}
\Large
\captionsetup{font=Large}
\caption{Joint Training of All Agents by PPO}
\label{a.alg2}
\textbf{Symbols and Notations}:\\
    $N_{iter}$: number of iterations for the algorithm\\
    $N_{workers}$: number of threads running in parallel\\
    $N_{epochs}$: number of training epochs\\
    $s^{(n)}_t$: all limbs' states of agent $n$ at time $t$\\
    $a^{(n)}_t$: all limbs' actions of agent $n$ at time $t$\\
    $r^{(n)}_t$: rewards of agent $n$ at time $t$\\
    $rb$: replay buffer shared by all robots.\\
    $P_n$: probability of sampling robot n\\
    $T_{episode}$: maximum time step of an episode
\begin{algorithmic}[1]
\STATE \textbf{Initialize:} init \textit{GCNT} with random parameter $\theta$, init P from a uniform distribution.
\FOR{$i = 1, \dots, N_{iter}$}
    \STATE // Collect robot experience in parallel
    \FOR{$j = 1, 2, \dots, N_{workers}$}
        \STATE Sample a robot $n \in N$ based on the probability $p_n$
        \STATE // Collect one episode of experience
        \FOR {t = 0, \dots, $T_{episode}$} 
            \STATE $a^{(n)}_t$ $\gets$ \textit{GCNT}($s^{(n)}_t$)
            \STATE $s^{(n)}_{t+1},r^{(n)}_t,done^{n}$ $\gets$ simulate($e,a^{(n)}_t$)
            \STATE add \{$s^{(n)}_t,s^{(n)}_{t+1},a^{(n)}_t,r^{(n)}_t,done^{n}$\} to $rb^{n}$
            \IF {$done^n$}
                \STATE break
            \ENDIF
        \ENDFOR
    \ENDFOR
    \STATE // Training
    \FOR{$j = 1, \dots, N_{epochs}$}
        \STATE \textit{GCNT} $\gets$ train with PPO(\textit{GCNT}, rb)
    \ENDFOR
    \STATE // Update sampling probability
    \IF{$i \geq N_{warmup}$}
        \STATE $P \gets \text{samplingProbUpdate}(R)$, depending on the episode length of each robot.
    \ENDIF
\ENDFOR
\end{algorithmic}
\end{algorithm}

\newpage

\section{Weisfeiler-Lehman Algorithm}

In this section, we present the detailed process of the original Weisfeiler-Lehman algorithm, with its pseudocode shown in Algorithm \ref{a.alg3}.

\begin{algorithm}
    \Large
    \captionsetup{font=Large}
    \caption{Weisfeiler-Lehman Algorithm}
    \label{a.alg3}
    \textbf{Input}:Graph $\mathcal{G}=\langle\mathcal{V}, \mathcal{E}\rangle$, node type\\
    \textbf{Output}:String of node labels
    \begin{algorithmic}[1]
    \STATE // First: Initialize:
    \STATE converged = False
    \FOR{each vertex $v$ in $\mathcal{V}$}
		\STATE label[$v$] = nodeType[$v$]
    \ENDFOR
    \STATE
    \STATE // Second: Iteratively update labels:
    \WHILE{not converged}
        \STATE new\_label = [ ]
        \FOR{each vertex $v$ in $\mathcal{V}$}
			\STATE // Collect labels of neighbors
			\STATE neighbor\_labels = [ ]
            \FOR{each vertex $u$ in neighbors($v$)}
				\STATE neighbor\_labels.append(label[$u$])
			\ENDFOR
        	\STATE // Sort neighbor labels and concatenate with current label
        	\STATE new\_label = (label[$v$], sorted(neighbor\_labels))
        	\STATE // Hash the combined label to get a new label
        	\STATE new\_label[$v$] = hash(new\_label)
		\ENDFOR
        \STATE
        \STATE // Check for convergence
        \IF{new\_label == label}
            \STATE converged = True
        \ELSE
            \STATE label = new\_label
        \ENDIF
    \ENDWHILE
    \STATE
    \STATE // Third: Return final lables:
    \STATE \textbf{return} the string of node labels
    \end{algorithmic}
\end{algorithm}

\newpage

\section{Experimental results of GCNT using Hypernetwork.}

\begin{figure}[!htbp]
\centering
\includegraphics[width=1\textwidth]{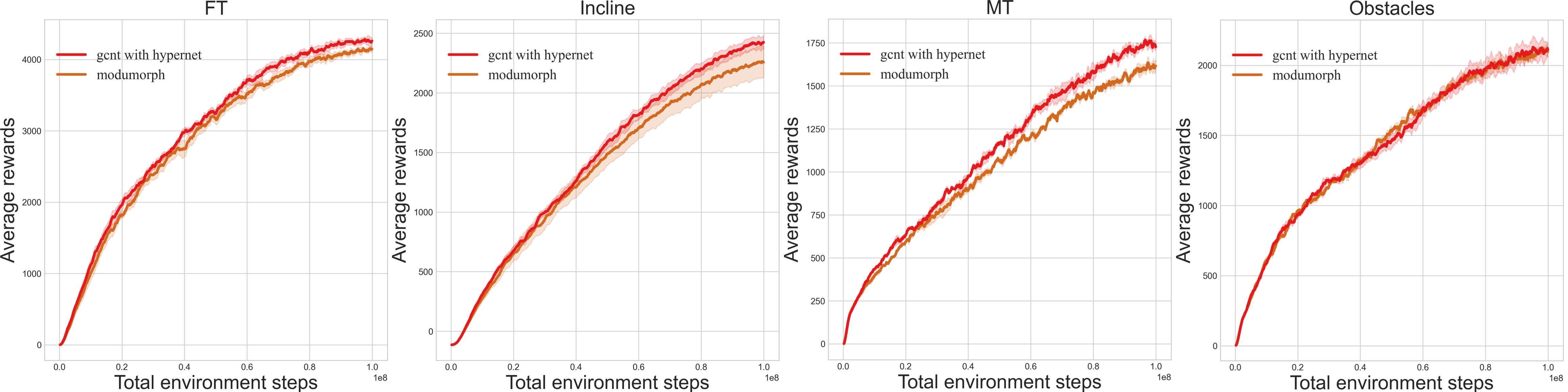}
\captionsetup{font=Large}
\caption{Average reward of GCNT in UNIMAL environment after using Hypernetwork.}
\label{a.fig1}
\end{figure}

As mentioned, ModuMorph uses a Hypernetwork to generate different encoders and decoders for each robot limb. This improves the performance of the network to some extent. However, the Hypernetwork introduces a larger number of parameters. The number of parameters in ModuMorph is 2-3 times that of our method, GCNT. In contrast, our method has fewer parameters but achieves similar performance to ModuMorph. More importantly, our method can also integrate a Hypernetwork to generate different encoders and decoders for each limb. Figure \ref{a.fig1} shows the average reward curve of GCNT after using the hypernetwork on different environments of UNIMAL. It can be seen that GCNT with Hypernetwork performs better than ModuMoroh. We believe that this results from our method being able to better extract and utilize morphological information.\\

\section{Index inconsistency issue caused by depth-first traversal}

\begin{figure}[!htbp]
\centering
\includegraphics[width=1\textwidth]{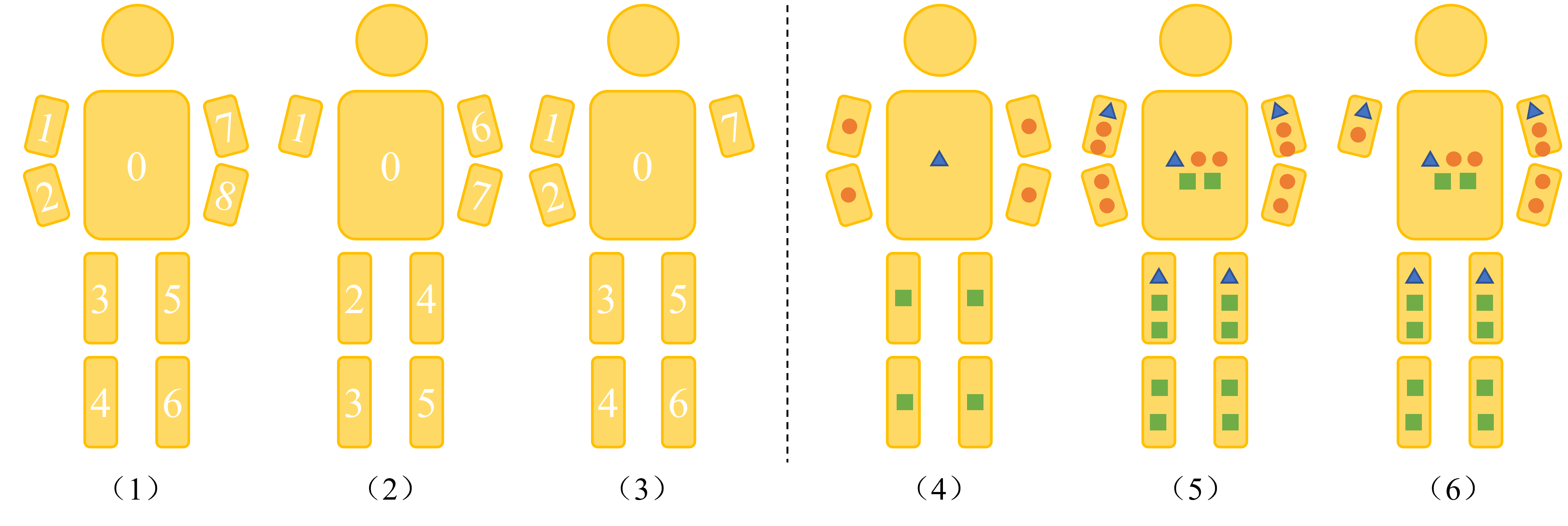}
\captionsetup{font=Large}
\caption{Comparison of methods based on depth-first traversal and convergence.}
\label{a.fig2}
\end{figure}

The left side of Figure \ref{a.fig2} shows the indexing inconsistency issue caused by the depth-first traversal method, where the same type of limb has different indexes in different robots. (1) and (2) differ by only one arm, and (3) can be considered as a horizontal flip of (2). They have almost identical shapes, but their limb indexes are significantly different. This makes it difficult for the control strategy learned based on indexes to transfer between different morphologies. For example, the strategy learned on the thigh in (2) (indexes 2 and 4) cannot be directly transferred to the thigh in (3) (indexes 3 and 5). In contrast, the right side of Figure \ref{a.fig2} shows a method based on aggregation. (4) represents the initial state of the robot, where different types of limbs have different attributes (circle, triangle, square). During convergence, each limb receives information from its neighbor limbs. (5) and (6) show the state of two robots with different morphologies after converging once, respectively. It can be seen that the properties of the same type of limbs maintain great similarity. This is more conducive to the transfer of learned strategies. Therefore, we believe that methods based on aggregation, such as GCN, have better generalization.\\

\section{The number of parameters in the networks.}

Here, we report the number of parameters in GCNT and the baselines. In SMPENV, all environments have the same observation space, so the number of parameters remain the same across environments. In UNIMAL, the MT and Obstacles environments provide additional observations (height map), so the algorithm's parameter count is no longer uniform. Therefore, we report the parameter count for each method in each environment separately. The details are shown in Table 7 and Table 8.

\begin{table}[htbp!]
\Large
\centering
\begin{tabular}{ccccc}
\hline
\textbf{}       & \textbf{SMP} & \textbf{AMORPHEUS} & \textbf{SWAT}   & \textbf{GCNT(Ours)} \\ \hline
\textbf{Params} & 694410       & 202389             & 202397          & 206613              \\ \hline
\end{tabular}
\captionsetup{font=Large}
\caption{The number of parameters for each method in SMPENV.}
\label{a.tab7}
\end{table}

\begin{table}[htbp!]
\Large
\centering
\begin{tabular}{ccccc}
\hline
\textbf{} & \textbf{MetaMorph} & \textbf{MetaMorph*} & \textbf{ModuMorph} & \textbf{GCNT(Ours)} \\ \hline
\textbf{FT}        & 3318147            & 3315075             & 5232774            & 3333583             \\
\textbf{Incline}   & 3318147            & 3315075             & 5232774            & 3333583             \\
\textbf{MT}        & 3531587            & 3528515             & 8608262            & 3553523             \\
\textbf{Obstacles} & 3539907            & 3536835             & 9689862            & 3561843             \\ \hline
\end{tabular}
\captionsetup{font=Large}
\caption{The number of parameters for each method in UNIMAL.}
\label{a.tab8}
\end{table}

\section{Behavior analysis}

\begin{figure}[!htbp]
\centering
\includegraphics[width=1\textwidth]{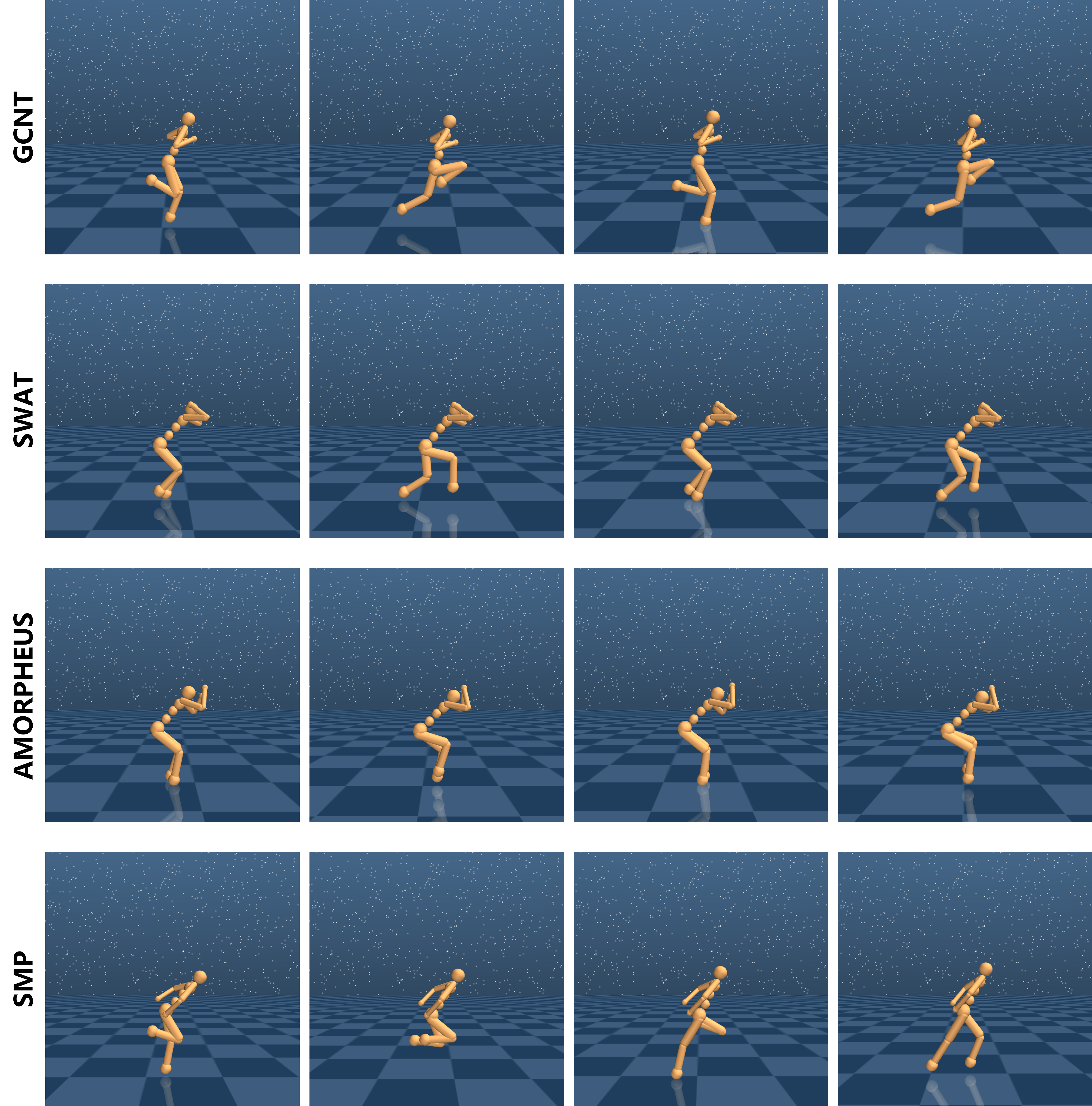}
\captionsetup{font=Large}
\caption{The trajectory of humanoid\_2d\_9\_full in CWHH++.}
\label{a.fig3}
\end{figure}

In this section, to compare the control performance of the universal policy network obtained by different methods, we apply them to humanoid\_2d\_9\_full in the CWHH++ environment and visualize its movement trajectory. Details are shown in Figure \ref{a.fig3}. It can be seen that GCNT achieved the best performance because of the effective use of the robot's morphological information. SWAT did not sufficiently extract morphological information, resulting in smaller stride lengths and slower motion compared to GCNT. AMORPHEUS did not use the robot's morphological information. The agent does not walk by alternating its legs, but jumps forward like Hopper. SMP moves very unsteadily and the movements on the left and right legs are not symmetric. We believe that this is caused by the loss of transmitted messages.

\end{document}